# Graph Neural Network with One-side Edge Sampling for Fraud Detection

Hoang Hiep Trieu

*Abstract—* **Financial fraud is always a major problem in the field of finance, as it can cause significant consequences. As a result, many approaches have been designed to detect it, and lately, Graph Neural Networks (GNNs) have been demonstrated as a competent candidate. However, when trained with a large amount of data, they are slow and computationally demanding. In addition, GNNs may need a deep architecture to detect complex fraud patterns, but doing so may make they suffer from problems like over-fitting or over-smoothing. Over-fitting leads to the reduced generalisation of the model on unseen data, while over-smoothing causes all nodes' features to converge to a fixed point due to the excessive aggregation of information from the neighbour nodes. In this research, I propose an approach called One-Side Edge Sampling (OES) that can potentially reduce the training duration as well as the effect of over-smoothing and over-fitting. The approach leverages the predictive confidence in edge classification task to sample edges from the input graph during a certain number of epochs. To explain why OES can alleviate over-smoothing, I will perform a theoretical analysis of the proposed approach. In addition, to validate the effect of OES, I will conduct experiments on different GNNs on two datasets. The results show that OES can empirically outperform backbone models in both shallow and deep architectures while also reducing training time.**

## I. Introduction

Financial fraud is always a major problem for the economy. 1.17 billion pounds has been lost by fraudsters (UK Finance Limited, 2025), and the number of fraud incidents has increased by 33% from last year, reaching 4.1 million (Office for National Statistics, 2025). This indicates that financial fraud is and will continue to cause severe consequences, and as a result, it is urgent to have a system to detect fraud effectively.

Historically, traditional methods like rule-based or traditional machine learning algorithms have been applied to deal with the problem, but those have been proven to be ineffective (Sahin and Duman, 2011; Seeja and Zareapoor, 2014). In contrast, Graph Neural Network (GNNs) have shown excellent results on the task (Liu et al., 2021; Cardoso et al. 2022; Yoo et al., 2023; Cheng et al., 2024). One advantage of GNNs is that they can analyse the network structure and discover patterns that machine learning or rule-based approaches may not be able to. In addition, GNNs can provide robust inductive biases, using graphs created by modelling accounts and transactions as nodes and edges as the input, allowing more in-depth pattern recognition and relational inference. Furthermore, GNNs allow deep learning methods to be applied to graph data, hence increasing the accuracy of the predictions. These characteristics make GNNs well-equipped to detect complex fraudulent patterns.

However, there are some challenges in building effective GNNs that need to be addressed. First, training time can be long and computationally demanding. GNNs require all data to be transformed into graphs, making the computation process sparse, therefore difficult to perform efficiently (Serafini, 2021). The problem becomes more severe in fraud detection in banking, as there is always a large number of transactions that need to be processed. There are some suggestions to reduce the training time, either by applying frameworks for distributed training (Wang et al., 2019; Fey and Lenssen, 2019) or by reducing the scale of the graph (Zeng et al., 2020; Zhao et al., 2023). Nevertheless, those applications are either not designed for GNNs or will reduce the performance as a trade-off. In addition, over-fitting and over-smoothing are other crucial concerns. Over-smoothing happens when the node representations are over-processed to the point that they are all similar. This will negatively affect the performance of the GNNs since they will not be able to use the information from the node's features. Over-fitting happens as the model is unable to generalise to the test data but is capable of learning the training data well. Both problems have gained significant attention due to their negative effect on GNNs' performance in general and deep GNNs in particular. It is important to build GNNs with a deep architecture since they can detect complex money laundering patterns that shallow GNNs cannot. For example, a 2-layer GNN can only extract information from at most 2-hop neighbourhoods. As a result, a complex fraud cycle where the money is transferred through many different accounts may not be detected. Some approaches have been designed to address them (Xu et al., 2018; Rong et al., 2020; Zhou et al., 2021; Rusch et al., 2022). However, those algorithms are either not guaranteed to alleviate over-smoothing and over-fitting or face the risk of sacrificing predictive power in return.

I suggest One-side Edge Sampling (OES), an edge-sampling technique that can improve the scalability of GNNs and alleviate over-fitting and over-smoothing. The main idea of the approach is to leverage the predictive confidence of the edge classification task to guide the edge-sampling process. Specifically, I measure how confident each edge prediction is, then filter the edges that have their predictive confidence higher than a threshold and whose label is the same as the ground truth. After that, at a certain number of epochs, I will sample them with a fixed ratio and create a new graph that will be used during the training process.

The contribution of the research can be summarised as follows:

(1) I propose a novel edge-sampling technique, OES, that takes into account the confidence of edge prediction to guide the sampling process.

(2) I show why applying OES can alleviate over-smoothing in theory.

(3) I validate the effect of OES on the IBM Anti-Money Laundering dataset with three GNN backbones of different numbers of layers. The experiments demonstrate that OES can reduce training duration as well as over-fitting and over-smoothing.

(4) My method can easily be integrated into different GNN backbones as the method uses the edge prediction, which can be generated by many GNN models.

The rest of the research is structured as follows: Chapter 2 will cover GNNs and their recent advancements on fraud detection, as well as their problems related to scalability, over-smoothing, and over-fitting. In Chapter 3, I will describe the backbone models as well as the proposed method and show how it can alleviate over-smoothing and over-fitting. Chapter 4 will cover how the experiment is conducted and Chapter 5 will focus on evaluating the project outcome with an explanation of the effect of OES on over-smoothing and over-fitting, as well as a discussion of parameter sensitivity. Chapter 6 will present a conclusion on the findings of the project, with suggestions for future work.

## II. RELATED WORK

### A. Graph Neural Network in fraud detection

Overall, there are two main classes of GNNs: spectral-based and spatial-based. Spectral-based GNNs involve the use of the Fourier transform to convert the node representation into the spectral domain. One of the aims of GNNs in this domain is to learn a good filter that is capable of capturing the important characteristics of the input graph. Specformer (Bo et al., 2023) uses an advanced filter by considering the relationship between eigenvalues, which contains important geometric information. GPR-GNN (Chien et al., 2020) suggests a polynomial filter, created by using a polynomial function, leveraging the PageRank information. FAGCN (Bo et al., 2021) simplifies the filter function by using a linear filter, combined with some predefined graph filters to improve scalability while still maintaining a high performance. Spatial-based GNNs are performed by using the message-passing mechanism on a graph, meaning the node's state is repeatedly updated using information from the neighbouring nodes. It is used more in fraud detection and represented by a more prominent family of GNNs known as Message Passing Neural Networks (MPNNs). One of the crucial works is Graph Convolutional Networks (GCN), the first to apply a convolutional layer to graph-structured data (Kipf and Welling, 2017). Then GATs introduce an attention mechanism where it learns the edge's weight adaptively, allowing it to calculate the node representation more accurately (Veličković et al., 2018). Another promising direction is related to the Weisfeiler-Lehman (WL) isomorphism test. For a long time, there has been no algorithm that can outperform that (Garey and Johnson, 2002; Babai, 2016). Xu et al. (Xu et al., 2018) have proven that MPNNs can be equally powerful as the WL test and suggested Graph Isomorphism Networks (GIN). Based on that, there are multiple approaches aiming to enhance the model, either by adding node IDs (Loukas, 2019), random node initialisation (Abboud et al., 2020), local graph counts (Barceló et al., 2021; Bouritsas et al., 2022), positional node embeddings (Dwivedi et al., 2021; Egressy and Wattenhofer, 2022) or heterogeneity factor (Wójcik, 2024).

Recently, GNNs have been proven to be one of the best tools to detect fraud. In credit card fraud detection, notable suggestions can include designing a weighted multiple graphs to capture dynamic changes in the transaction network (Liu et al., 2021) or using the spatial-temporal attention mechanism to increase performance. (Cheng et al., 2022). In addition, Nguyen et al (Nguyen et al., 2022) leverage future information after the transaction is done, and Xiang et al (Xiang et al., 2023) leverage the temporal graph and attention mechanism to enhance fraud detection ability. In insurance fraud, significant contributions can be named, like utilising non-fraud behaviour to recognise customer patterns (Zhang et al., 2022) or using a dual-level learning framework to learn node-level and super-node level representations (Zheng et al., 2023). GNNs are also effective in detecting money laundering. Some of the recent advancements in the field have successfully leveraged the bipartite graph to enhance GNNs' predictability. Cardoso et al have successfully leveraged a customer-transaction bipartite graph to improve the GNNs' predictability to detect unusual patterns (Cardoso et al., 2022). Having a similar idea, Rao et al also use a bipartite graph, but for the customer-product relationship, to detect credit card fraud (Rao et al., 2021). Other ideas, like (Liu et al., 2018) or (Lu, Tsai and Li, 2024), have also used different relationships in a bipartite graph to increase model performance. Some other advancements can be named like group-aware deep graph learning (Cheng et al., 2023), which can detect organised money laundering activities, or leveraging attention mechanisms to capture relationships between location and timeframe of transaction networks (Khosravi et al., 2025).

However, all these models are complex and require a lot of computational resources. Especially since the size of the transaction graph is always large due to the large volume of transactions, in combination with the need to develop models urgently in practice to keep up with the demand, it is difficult to implement those models in practice without a technique to reduce training time

### B. Graph sampling technique

There are some suggestions to improve GNNs' scalability by using a distributed system to train the model in parallel (Wang et al., 2019; Fey and Lenssen, 2019). Nevertheless, distributed training is not suitable for a graph since a graph is not a collection of independent observations but a single data structure consisting of interconnected nodes and edges. As a result, it is not possible to partition nodes alone since we will lose valuable information related to their connections.

Some other suggestions, based on sampling methods, have been proven to be more promising lately. Node-wise sampling has been suggested by GraphSAGE (Hamilton et al., 2017) and S-GCN (Chen et al., 2018), with the idea being only to conduct convolutional calculation on not all but some neighbours of nodes. Combined with mini-batch training, the computation time and memory requirements are reduced significantly. Another promising node-wise approach is ATP (Li et al., 2024), a method that takes into account a node's topological uniqueness to enhance other sampling methods. On a different direction, layer-wise sampling is suggested by FastGCN (Chen et al., 2018) and AS-GCN (Huang et al., 2018), adding a layer to sample nodes' neighbours. HDSGNN (Li et al., 2024) is another effective approach, using historical information of nodes' embedding to reduce the variance in performance caused by the sampling method. Those recent advancements in sampling methods have opened the opportunities to apply a deeper expansion of GNNs. However, all the above approaches have not addressed the problem of over-fitting and especially over-smoothing, the main difficulty of building a deep network (Li, Han and Wu, 2018).

*C. Over-fitting and over-smoothing*

On over-fitting, different methods have been suggested in order to mitigate the problem. Some approaches are related to the distribution and correlation of the data. StableGNN (Fan et al., 2021) leverage the graph high-level representations to detect spurious correlation to increase the generalisation of GNNs. L2R-GNN (Chen et al., 2024) improves the model generalisation ability by reducing correlations between variables of different clusters with a non-linear approach, while Choi et al (Choi et al., 2025) suggest an iterative learning strategy to adapt to the change in distribution of the dataset by shifting both the features and parameters. Other methods either focus on applying a regularisation technique (Yang et al., 2021) or label propagation (Wang and Leskovec, 2020). Even though all those approaches have achieved significant improvement, they do not address other concerns like scalability or over-smoothing.

Over-smoothing is introduced by Li et al (Li et al., 2018), then generalised further by Oono and Suzuki (Oono and Suzuki, 2020), and defined as a phenomenon when, as the layers of a GNN model increase, the features of the nodes will converge to a stationary point. As a result, all node features will be similar to each other. Even though it has been proven that some level of smoothing is beneficial for tasks like regression and classification (Keriven, 2022), an excessive level of that will result in a convergence to a non-informative value. This phenomenon will restrict the output of deep GNNs to be independent of their input node features, which significantly reduces the predictive power of the models since node features contain important information for many tasks, such as edge classification. Previous experiments have demonstrated that the performance of the GNN model will decrease because of over-smoothing when the number of layers is higher than three (Kipf and Welling, 2017). Historical approaches aiming to mitigate the problem can be categorised into three ideas. The first idea is to focus on designing the GNNs architecture. He et al (He et al., 2015) utilise the learning residual functions, while AGNN (Chen et al., 2023) combines graph convolutional and embedding layers, to overcome this problem. The second idea is to use a normalisation method, and previous works can include the use of batch (Li et al., 2019) and pair (Zhao and Akoglu, 2020) normalisation. The third idea is related to the data augmentation technique and has been able to achieve notable improvement lately due to being straightforward, efficient, and easy to implement in different architectures. Rong et al suggest DropEdge (Rong et al., 2020), an approach that samples edges in the graph randomly with a fixed rate to create a new subset of edges. However, the approach assumes that all edges are equally important. Therefore, DropEdge ignores other important information that can provide a more valuable guide. Some other approaches, such as introducing another neural network (Hasanzadeh et al., 2020) or a learnable parameter and line graph transformation (Morshed et al., 2023) to determine which edges to drop, also have a positive effect on dealing with the problem. However, the two above techniques might be too complicated, as they will increase the training time of the model

III. METHODOLOGY

In this chapter, I will introduce the baseline models and the methodology of One-side Edge Sampling. In addition, I will discuss how applying it will affect the over-smoothing and over-fitting problem

*A. Preliminaries*

Given a directed multigraph $G = (V, E)$, where $V$ represents a set of nodes, and $v \in V(G)$ represent accounts. $E$ is a set of edges and $e = (u, v) \in E(G)$ represent a directed transaction from $u$ to $v$. The adjacency matrix is denoted as $A$ and the size of edges is denoted as $|E|$. All features of $v$ are denoted as $h(v)$ and features of $e$ is denoted as $h(u, v)$. The incoming neighbours of $v$ is denoted as $N_{in}(v)$ and the outgoing neighbours is denoted as $N_{out}(v)$. The global attribute $g$ describes the overall structure and relationship between nodes and edges.

The message-passing mechanism will follow three steps iteratively:

(1) At iteration $k$, each node will send its current information $h^{k-1}(u)$ to its neighbours.
(2) Using that information, each node will aggregate that to create an embedding $a^k(v)$:
$$a^k(v) = \psi(\{h^{k-1}(u) | u \in N(v)\})$$
(3) Using the node's previous state $h^{k-1}(v)$ and new embedding $a^k(v)$ each node will update its information:
$$h^k(v) = \phi(h^{k-1}(v), a^k(v))$$

where $\psi$ is an aggregation function (*e.g.,* mean, max), $\phi$ is an update function (*e.g.,* linear combination), and $\{\}$ denotes a multiset since this is a multigraph. In our case, the mechanism is adjusted to adapt to a directed multigraph. For a directed graph, the aggregate step will only use the incoming information from neighbours:
$$a^k(v) = \psi(\{h^{k-1}(u) | u \in N_{in}(v)\})$$

And the edge feature of directed edge $e$ can also be considered:

$$a^k(v) = \psi(\{h^{k-1}(u), h^k(u,v) | u \in N_{in}(v)\})$$

To determine whether an edge is positive or negative, we utilise the generated embedding to get how likely an edge's label to be positive or negative:

$$y^{pos}, y^{neg} = \sigma(m(a(u), a(v)) | u \in N_{in}(v), v \in N_{out}(v))$$

where $m$ can be an inner product or multi-layer perceptron (MLP) and $\sigma$ can be a point-wise non-linear function. In this study, $m$ is MLP and $\sigma$ is Linear. Then, the label of the edge is decided by which label's value is higher:

$$\hat{y} = argmax(y^{pos}, y^{neg})$$

### B. Baseline models

In this part, I will introduce the baseline models of the experiment, consisting of GIN with edge features (Xu, K. et al., 2019), denoted as GIN, GIN with ego IDs (You et al., 2021) denoted as GIN+EGO and GIN with edge updates (Battaglia et al., 2018), denoted as GIN+EU.

#### 1) Graph Isomorphism Network

GIN has been proven to be one of the first algorithms to match the power of the Weisfeiler-Lehman (WL) test (Xu, K. et al., 2019). The purpose of the WL test is to determine whether two graphs are topologically identical. A graph labelling procedure is performed, where labels are iteratively assigned to different nodes based on the labels of neighbouring nodes. Two graphs can be considered isomorphic if their nodes' labels are similar at all iterations. It has been proven that the WL test, with enough iterations, can distinguish non-isomorphic graphs that otherwise cannot be done by any previous GNNs (Maron et al., 2019).

GIN can match that performance, and until now, it remains a state-of-the-art model because of its strong expressive power. GIN can capture and represent the patterns and dependencies in graph-structured data by leveraging the message-passing scheme to update the node representation. Specifically, the update step is changed to:

$$h^k(v) = \phi((1 + \varepsilon^k) * h^{k-1}(v) + \sum_{u \in N_{in}(v)} h^{k-1}(u))$$

where $\phi$ is MLP and $\varepsilon$ can be a learnable or fixed scalar. The MLP enable GIN to learn the injective functions for embeddings as it can help the algorithm learn effectively, because of the universal approximation theorem (Hornik, 1991). Then, the node embeddings learned by GIN can be directly used for edge classification, with the same mechanic as regular GNNs.

#### 2) Graph Isomorphism Network with Ego IDs

One of the important aspects of detecting fraud is cycle detection. A popular money laundering pattern is a circular flow, meaning the money is transferred from one account to many others before eventually returning to the original one. GIN+EGO (You et al., 2021) is suggested to detect those with the main approach of assigning a unique, binary, and distinct label to the centre nodes. For a given node $v$, the $k$-hop ego network $G_v^k$ is extracted, and the central node is assigned a label. As a result, this node can recognise cycles that it is a part of by recognising when a series of messages goes back around to it.

To be more specific, GIN+EGO first assigns a unique label to centre nodes. As a result, it divides nodes in the graph structure into two types: nodes with labels and nodes without. Those labels are utilised in the next step by the message-passing mechanism and applied to all the ego networks. However, instead of passing $h^{k-1}(u)$ to neighbour nodes, the message-passing function of GIN+EGO is:

$$m^k(u) = \theta_{[s=v]}(h^{k-1}(u))$$

where $s$ and $v$ denote the label of the node and its neighbour, $[s = v]$ indicates that 1 if $s = v$ else 0 and $\theta$ can be a variety of functions. Then, the aggregate function will utilise the output of the message-passing function to calculate the node embedding:

$$a^k(v) = \psi(\{m^k(u) | u \in N(v)\})$$

By adding the indicator, there will be two sets of message-passing function: $\theta_1$ for labelled nodes and $\theta_0$ for unlabelled ones. The passing function can also be extended to using edge features:

$$m^k(u) = \theta_{1[s=v]}(h^{k-1}(u), h^{k-1}(u,v))$$

#### 3) Graph Isomorphism Network with edge updates

The main idea of GIN+EU is to integrate a Graph Networks (GN) block into its architecture. GN block is a flexible and easy-to-adapt structure that can transform graphs. Given an input graph $G$, the function will return an isomorphic graph $G'$, with updated edges, nodes, and global feature. The main idea of the GN block is to iteratively update the graph, from edge to node and then to the global level, between the training epochs. The computation steps of the GN block are described as follows:

(1) Each edge feature is updated, taking into account its feature, the receiver and sender nodes' features and the global attribute:
$$h'(u,v) = \phi(h(u,v), h(u), h(v), g | u \in N_{in}(u), v \in N_{out}(u))$$

(2) Then, using the previous outputs, an aggregation is created to use for nodes' update:
$$\bar{h}'(u,v) = \psi(h'(u,v))$$

(3) After that, the computation moves to the node level, where the node feature is updated based on the previous aggregation and its node feature:
$$h'(v) = \phi(\bar{h}'(u,v), h(u), h(v), g | u \in N_{in}(u), v \in N_{out}(u)))$$

(4) Then, all edges' updates are aggregated to use in the global attribute update:
$$a(u,v) = \psi(\bar{h}'(u,v))$$

(5) Similarly, all nodes' updates are aggregated to use for the global attribute update:
$$a(v) = \psi(h'(v))$$

(6) Finally, the global attribute is updated based on the previous two above information:
$$g' = \phi(a(u,v), a(v), g)$$
where $\psi$ can be a variety of permutation invariant functions like mean or max and $\phi$ can be Linear activation function, MLP or Recurrent Neural Network (RNN). Then a new graph $G'$ is created with updated components after going through the GN block.

Adding a GN block in a GNN's architecture can help the model to learn the relational inductive bias within the graph (Battaglia et al., 2018). Relational inductive bias refers to the assumption that the relationships between nodes are vital for learning. For example, the assumption that there is a fraudulent transaction between two nodes can be expressed by edges between those nodes. GNN with the GN block can determine those relationships by considering a wider range of aspects, such as global attributes. In addition, the GN block is flexible as its output can be adapted to different tasks or different graph structures. In this study, the GN block is tailored to be edge-focused, meaning using edges as output, since this is an edge classification task.

### C. Proposed method

OES has three parameters: $p, r$, and $n$, where $p$ denotes the percentile threshold, $r$ denotes the sampling ratio, and $n$ denotes the number of epochs that utilised the method. The core of OES is to sample edges that the model can predict correctly with confidence. The predictive confidence of an edge, from $u$ to $v$, denoted as $c(e_{u,v})$, is measured by the highest value of the model's output:
$$c(e_{u,v}) = max(y_{u,v}^{pos}, y_{u,v}^{neg}) \quad (1)$$
This measurement represents a guide to the process. I use a set of $c(e_{u,v})$, denoted as $C$, to calculate the threshold, denoted as $P_p$, as follows:
$$P_p = Percentile(C, p) \quad (2)$$
At a certain number of training epochs, OES samples edges that has $c(e_{u,v})$ higher than $P_p$, and its prediction is similar to the ground truth with a certain ratio. After that, a list of edges is filtered based on its accuracy and predictive confidence :
$$E' = \{e_{u,v} | c(e_{u,v}) \geq P_p \text{ and } \hat{y}_{u,v} = y_{u,v}\} \quad (3)$$
At each epoch where the method is used, the rate of dropped edges will be $(1-p) * r$. Formally, OES creates a new set of edges $E_{OES}$, with a reduced edge size $|E_{OES}| = (1-p) * r * |E|$. Then a new graph $G_{OES} = (V, E_{OES})$ is created and replaces the initial graph $G$ during the training process.

OES will stop after a number of epochs $n$ because, at a certain point, dropping edges will only have a marginal effect on training time and will reduce predictability. The method is only utilised in the training process, not in validation and testing. The process is described in Algorithm 1.

---

**Algorithm 1**: One-side Edge Sampling
**Input**: Directed multigraph $G = (V, E)$, percentile threshold $p$, sampling ratio $r$, number of epochs $n$, ground truth $y$, predictions $y_{pos}, y_{neg}$
**Output**: New graph $G_{OES}$
1: *if* epoch $\leq n$ *do*
2:     Obtain $c(e_{u,v})$ with Eq. (1);
3:     Create set $C = \{c(e_{u,v})\}$;
4:     Calculate $P_p$ with Eq. (2);
5:     Obtain $E'$ with Ep.(3);
6:     Select m = $|E'| * r$ edges and create a new set $E_{OES}$;
7:     Create new graph $G_{OES} = (V, E_{OES})$;
8: *end if*
9: *return* $G_{OES}$

---

### D. On preventing over-fitting

Traditionally, the over-fitting problem can be reduced by a data augmentation technique. By introducing randomness into the input data, the model will not memorise the input data but will be able to generalise better. Different methods, such as Dropout (Srivastava et al., 2014) for Neural Networks or Cutout (Devries and Taylor, 2017) and RandomErasing (Zhong et al., 2020) for Convolutional Neural Networks, are built based on this approach.

OES can have the same effect as a data augmentation technique. It modifies the edge connections and, as a result, creates a new version of the input graph. As a result, it forces the model to use that new version of the graph instead of the full graph during the training process. Hence, the negative effect of over-fitting can be reduced to some extent.

### E. On preventing over-smoothing

In this part, I will provide a theoretical analysis, inspired by the previous work of Rong et al to explain that OES can alleviate over-smoothing, I will begin with the definitions of subspace and relaxed $\varepsilon$-smoothing layer, as well as one auxiliary lemma, from Rong et al (Rong et al., 2020):

*Definition* 1 (subspace): Let $\mathcal{M} = \{EC | C \in \mathbb{R}^{M \times C}\}$ be an M-dimensional subspace in $\mathbb{R}^{N \times C}$, where $E \in \mathbb{R}^{N \times M}$, $E^T E = I_M$ and M≤N (Rong et al., 2020)

*Definition* 2 (relaxed $\varepsilon$-smoothing): Given the subspace $\mathcal{M}$ and $\varepsilon (\varepsilon > 0)$, $\hat{l}(\mathcal{M}, \varepsilon) = \left\lceil \frac{\log \frac{\varepsilon}{d_\mathcal{M}(X)}}{\log(s\lambda)} \right\rceil$ is the relaxed smoothing layer, where $\lceil \cdot \rceil$ compute the ceil of the input, $s$ denotes the supremum of the filters' singular values over all layers, $d_\mathcal{M}(\cdot)$ is the distance between the input matrix and the subspace $\mathcal{M}$, and $\lambda$ denotes the second largest eigenvalue of $A$ (Rong et al., 2020).

*Lemma* 1: The connection between $\lambda$ and $R_{st}$ for each connected component can be described as the following inequality (Rong et al., 2020):

$$\lambda \geq 1 - \frac{1}{R_{st}}(\frac{1}{d_s} + \frac{1}{d_t})$$

where $R_{st}$ denotes the total resistance between nodes $s$ and $t$, while $d_s$ and $d_t$ are the degrees of nodes $s$ and $t$, respectively.

$\hat{l}(M, \varepsilon)$ is proven to be approximately the threshold for the number of GNN layers before over-smoothing occurs (Rong et al., 2020). As a result, that threshold needs to be increased so a deeper GNN can be built without facing the problem. Another characteristic of over-smoothing is the aggregation to a fixed point of all nodes' features, leading to the loss of important information. The information loss is measured by the distance between the original space and its subspace, denoted as $N - M$. It is desired that the information loss value be decreased so we can retain more valuable information. I will provide the theoretical analysis to show that OES can alleviate over-smoothing by being able to increase $\hat{l}(M, \varepsilon)$ and decrease $N - M$, as follows:

*Theorem*: Denoting the original graph and its subspace as $G$ and $M$, and the new graph after applying OES and its subspace and $G_{oes}$ and $M_{oes}$, the following inequalities will hold:
1) OES is able to increase the relaxed $\varepsilon$-smoothing: $\hat{l}(M, \varepsilon) \leq \hat{l}(M_{oes}, \varepsilon)$.
2) OES is able to decrease the information loss caused by over-smoothing: $N - \dim(M) \geq N - \dim(M_{oes})$.

*Proof*: Removing edges from graph will only increase $R_{st}$ (Lovász et al., 1993). As a result, enough edges getting removed will make $R_{st} = \infty$, causing $\lambda$ to decrease and eventually equal to one. In addition, based on the formula of $\hat{l}(M, \varepsilon)$, we can see that $\lambda$ has an opposite effect on $\hat{l}(M, \varepsilon)$, meaning that when $\lambda$ decrease, $\hat{l}(M, \varepsilon)$ will increase. Therefore, reducing number of edges will cause $\hat{l}(\mathcal{M}, \varepsilon)$ to increase and we can conclude the first inequality. In addition, when $R_{st} = \infty$, the value of M will increase by one as when enough edges are dropped, the component will separate into two parts. Hence, $N - M$ will decrease and the second inequality is proven.

In conclusion, applying OES not only can increase the threshold for over-smoothing but also reduce the information loss caused by it, hence reducing the effect of over-smoothing.

## IV. EXPERIMENT

In this chapter, I will present the numerical experiments on the proposed approach and compare it against backbone models. The details will include the dataset, the preparation step, all backbone models, the evaluation metric, and the deployment procedures to help others reproduce the experiment. After that, I will try the approach on different depths to test its effectiveness on higher-depth GNNs. Lastly, a sensitivity analysis is performed to examine how different parameter values affect OES performance.

### A. Dataset

To test the impact of OES, I adopt the original experiment setting (Béni Egressy et al., 2024) on the IBM Anti-Money Laundering dataset (Altman et al., 2023). Specifically, I will use two versions of the small data, one has a high illicit ratio (HI) and one has a low illicit ratio (LI), denoted as HI_Small and LI_Small, respectively. The dataset is artificial, which means it was generated by simulations to mimic the behaviour and pattern of money laundering in real life. Dataset sizes and illicit ratios are shown in the Appendix.

### B. 4.2 Data preparation

#### 1) 4.2.1 Data transformation

A directed multigraph is created from the dataset, with nodes representing accounts and edges representing transactions. Each node will have a set of features, including account number and account balance. Each edge also has features like timestamp, transaction amount and currency. All edges will also have a label, based on the "Is Laundering" column. The transaction network is a directed multigraph since each transaction has a direction (from sender to receiver), and there can be multiple transactions between 2 accounts, which makes it a multigraph.

#### 2) Data splitting

First, the data is ordered by its timestamp. Then, the data is split into three sets: train, validation, and test, based on 2 timestamps $t_1$ and $t_2$ ($t_1 < t_2$) with the ratio of 60/20/20. The train set will be all transactions happened before timestamp $t_1$, the valid set will be between $t_1$ and $t_2$, while the test set will consist of all after $t_2$. Then all the set's data will be transformed into a graph. However, the training transactions must be accessible to the validation and test graphs in order to recognise previous patterns. Hence, while the train graph will only contain the train transactions, the validation graph will contain both training and validation ones, and the test graph will contain transactions from all three sets. However, only the transactions that are in the validation's time frame can be used for evaluation in the validation set, and only those in the test's time frame are used for the test set. This is similar to the situation that banks and other financial authorities faced, as they will have access to all transaction history.

### C. 4.3 Model evaluation

The problem of edge classification can be described as binary classification, where a positive label indicates a fraudulent transaction. Hence, the F1 score will be used for evaluation. The score measures the balance between precision and recall, meaning it can reflect the predictive ability comprehensively. The metric takes into account both false positives and negatives. Hence, it suits the characteristics of the dataset since it is very imbalanced. That characteristic requires a metric that can take into account both false negatives and positives. Other metrics like accuracy might not be useful here. This is also similar to what banks use in real-world scenarios. The training duration is also reported in minutes to reflect the scalability of models with and without OES. The formula of F1 can be described as:

$$F1 = 2 * \frac{Precision * Recall}{Precision + Recall}$$

where *Precision* will indicate the percentage of correctly predicted positive edges out of all positive prediction and *Recall* will show the many percentages of positive edge we can correctly cover with our prediction.

Each model is run 5 times with 5 different seeds, and the result is reported as the mean of the performance in each seed.

### D. Deployment

All models are trained using NVIDIA RTX A4000. I utilised implementation from PyTorch Geometric along with the original paper's GitHub source. The parameters I use in the experiment are as follows: percentile threshold: 99, sampling ratio: 10, number of epochs: 20. All other parameters of the backbone model, with and without OES, are kept the same for fair evaluation.

For evaluating the effect of OES on over-fitting and over-smoothing, the number of layers of GIN will be increased from 2 to 16. The original architecture of GIN is shallow. Hence, it is difficult to show the effect of OES because over-fitting and over-smoothing can be seen most clearly when the architecture is deep. Therefore, the number of layers will be increased to demonstrate the effect of OES more clearly.

A parameter sensitivity analysis is conducted on GIN on HI_Small data. In each experiment, one parameter is changed, and the other two are kept the same.

## V. DISCUSSION

### A. Result and Analysis

Table 1 summarises the performance of each model with and without One-side sampling. It can be seen that the effectiveness of OES is proven as the proposed approach consistently improves the performance of all backbone models. Noticeably, the performance of GIN+EU increases by nearly 10%. In addition, as a bonus, applying OES makes the models more consistent, as the standard deviation of all experiments, except for GIN on HI_Small data, is decreased. Table 2 shows the training time of each model. Algorithms with OES are consistently trained faster than the default versions. This indicates the positive effect of OES on reducing training duration.

**Table 1 Test F1-score (%) of different backbone models with and without OES**

|  | GIN | | GIN+EU | | GIN+EGO | |
|---|---|---|---|---|---|---|
|  | Original | OES | Original | OES | Original | OES |
| HI_Small | 33.42 ±4.48 | **37.35** ±**5.44** | 50.96 ±6.95 | **59.47** ±**4.47** | 35.32 ±3.27 | **35.35** ±**2.04** |
| LI_Small | 10.36 ±2.69 | **10.38** ±**2.48** | 15.76 ±11.53 | **19.52** ±**3.29** | 12.20 ±5.17 | **12.41** ±**3.68** |

**Table 2 Training time (minutes) of different backbone models with and without OES**

|  | GIN | | GIN+EU | | GIN+EGO | |
|---|---|---|---|---|---|---|
|  | Original | OES | Original | OES | Original | OES |
| HI_Small | 99.63 | **91.61** | 232.90 | **189.94** | 134.24 | **113.52** |
| LI_Small | 149.72 | **126.27** | 214,27 | **199,68** | 194.76 | **162.52** |

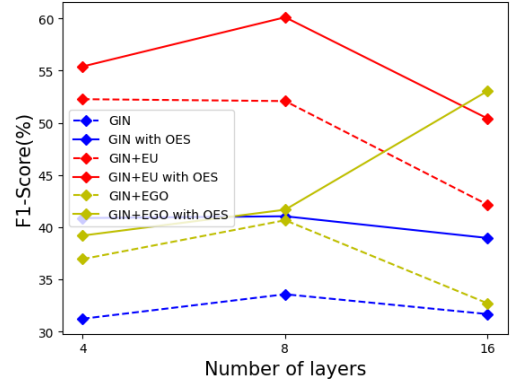

**Figure 1 Comparison of models' performance with and without OES in different number of layers**

Figure 1 shows the performance of the original models and the ones with OES at different depths. It can be seen that OES consistently improve the performance of backbone models throughout different numbers of layers. The difference in performance can be observed consistently throughout different numbers of layers, showing the effectiveness of OES at various depths. Especially, the biggest gap happened in the highest depth, showing that OES can still be effective with a high number of layers.

### B. Effect on preventing over-smoothing and over-fitting

From Figure 2, we can see that there is no over-fitting and over-smoothing problem. The backbone model, GIN with 16 layers, is shown to have a high test error while the train error is low. In addition, it is false to converge. On the other hand, when adding OES, the model not only obtains a lower train error, but the test error is lower and more stable. In addition, OES can make the model converge better. This indicates that OES is an effective tool to alleviate those problems. In addition, applying OES can help the model learn more meaningful node representations, as the train error is lower than that of the model without it. However, the effect of OES on GIN+EU and GIN+EGO with 16 layers is negligible. The reason is probably that these two algorithms implement a more sophisticated learning mechanism, either by adding a GN block or an MLP to the message-passing mechanism, thereby introducing an additional level of complexity that the current parameter setting of OES is insufficient to handle, leading to the problem of over-smoothing still happening.

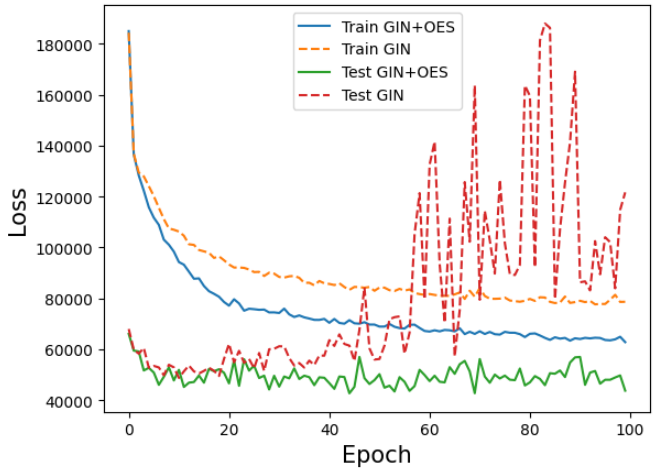

**Figure 2 Train and test loss comparison on GIN with and without OES**

*C. Sensitivity Analysis*

As the approach has 3 parameters, it is important to analyse the impact of each parameter on the model's performance. To test their impact, one parameter's value is changed at a time while the other two are equal to the default value. Table 3 presents the list of values for each parameter.

**Table 3 Parameter's description, symbol, range, and default value**

| Parameter | Description | Range | Default value |
|---|---|---|---|
| Percentile ($p$) | Percentile threshold to filter edges (%) | [99,95,90,80] | 99 |
| Sample ratio ($r$) | Ratio of sampling (%) | [20,10,5,1] | 10 |
| Number of epochs ($n$) | Number of epochs that execute OES | [10,20,30,50] | 20 |

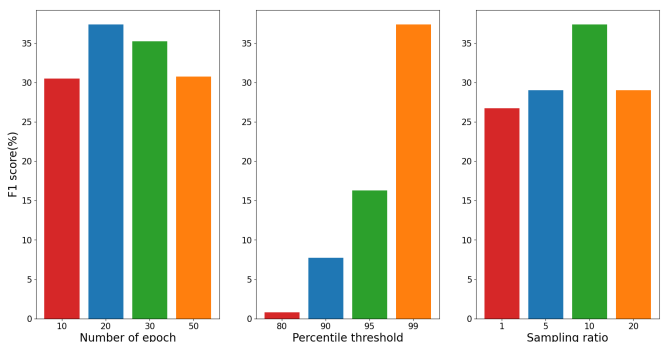

**Figure 3 Comparison of performance on different parameters**

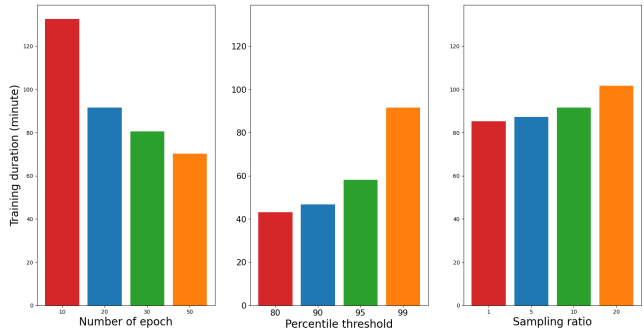

**Figure 4 Comparison of training time on different parameters**

Figures 3 and 4 show a comparison of different parameter values on the F1 score and training time. Overall, it can be seen that setting the right parameter can be crucial to the mode's performance since there is a wide gap between the performance of each parameter. The most distinguishable difference in performance can be seen in the percentile threshold, as the performance decreases drastically when the threshold is lowered, reaching nearly 0% at 80. The reason for the difference between each value of percentile threshold can be explained by its impact on the number of edges. It directly samples the original set of edges of the input graph, while the sampling ratio only relates to the already-sampled edges after being filtered by a percentile threshold. Therefore, a small change in $p$ can cause a big difference in the number of edges, hence heavily impact the model's performance. The other two parameters also have a sizeable impact on performance as the difference between the lowest and highest score is 5-10%. In addition, we can also see that it is necessary to find the right balance between removing too many or too few edges. For example, a too low value of $n$ or $r$ will reduce the impact of OES, making it less efficient to alleviate over-smoothing and over-fitting, leading to a lower performance score. As a result, the parameter needs to be tuned carefully to avoid the problem. On reducing training time, as the number of edges is dropped, the training process is faster. However, the effect is not linear. This can be explained by the fact that OES only affect the training process, not the validation and test.

## VI. CONCLUSION

In this research, I have proposed a novel edge-sampling method called OES. The method makes use of the predictive confidence of edge prediction to guide the sampling process. In a certain number of training epochs, by removing edges that have higher predictive confidence than a threshold and have correct predictions, the method can reduce the training time as well as the effect of over-fitting and over-smoothing. The theoretical analysis has shown OES effectiveness. In addition, the experimental results have empirically demonstrated that OES can improve the performance of GNNs.

Due to constraints on computer resources, the experiments cannot be conducted on a more complex model or larger data. In addition, the effect of OES on GIN+EU and GIN+EGO is still not clear. For future research, I will try to apply OES on different GNN architectures and datasets to further examine its effect, as well as fine-tune the parameters more to increase its effect on complex algorithms like GIN+EU and GIN+EGO.

ACKNOWLEDGMENT

I would like to thank Dr. Xiaowei Gu for his helpful comments and suggestions.